%% file: main.tex
\documentclass[sigplan,nonacm]{acmart}

\renewcommand\footnotetextcopyrightpermission[1]{}
\usepackage{booktabs}
\usepackage{multirow}
\usepackage{subcaption}
\usepackage{xspace}
\usepackage{graphicx}
\usepackage{amsmath}
\usepackage{enumitem}
\usepackage{url}
\usepackage{tikz}
\usepackage{xcolor}
\usepackage{algorithm}
\usepackage{algpseudocode}
\usepackage{tabularx}
\usetikzlibrary{arrows.meta, positioning, fit}

\newcommand{\tps}{tokens/s\xspace}

\newcommand{\sys}{PALS\xspace}

\begin{document}

\title{Power-Aware Runtime for Energy-Efficient LLM Inference}
\title{PALS: Power-Aware LLM Serving for Mixture-of-Experts Models}

\author{Can Hankendi}
\affiliation{%
  \institution{Boston University}
  \department{Electrical and Computer Engineering Department}
  \city{Boston}
  \state{MA}
  \country{USA}
}
\email{hankendi@bu.edu}

\author{Rana Shahout}
\affiliation{%
  \institution{Harvard University}
  \department{School of Engineering and Applied Sciences}
  \city{Cambridge}
  \state{MA}
  \country{USA}
}
\email{ranash.cs@gmail.com}

\author{Minlan Yu}
\affiliation{%
  \institution{Harvard University}
  \department{School of Engineering and Applied Sciences}
  \city{Cambridge}
  \state{MA}
  \country{USA}
}
\email{minlanyu@g.harvard.edu}

\author{Ayse K. Coskun}
\affiliation{%
  \institution{Boston University}
  \department{Electrical and Computer Engineering Department}
  \city{Boston}
  \state{MA}
  \country{USA}
}
\email{acoskun@bu.edu}

\begin{abstract}
Large language model (LLM) inference has become a dominant workload in modern data centers, driving significant GPU utilization and energy consumption. While prior systems optimize throughput and latency by batching, scheduling, and parallelism, they largely treat GPU power as a static constraint rather than a controllable resource. 

In this paper, we present a power-aware runtime for LLM serving, PALS, that treats GPU power caps as a first-class control knob and jointly optimizes them with software parameters such as batch size. The system combines lightweight offline power–performance models with a feedback-driven controller to select configurations that satisfy throughput targets while maximizing energy efficiency.

We implement PALS within an existing LLM serving framework, vLLM, demonstrating that it requires no model retraining or API changes. Across multi-GPU systems and both dense and mixture-of-experts (MoE) models, PALS improves energy efficiency by up to 26.3\%, reduces QoS violations by 4×–7× under power constraints, and tracks dynamic power budgets. These results highlight the potential of integrating power control directly into LLM inference runtimes, enabling energy-proportional and grid-interactive AI systems. 

\end{abstract}

\maketitle

\input{sections/01-introduction}

\input{sections/03-motivation}

\input{sections/new_method}

\input{sections/08-evaluation}
\input{sections/09-results}
\input{sections/10-discussion}
\input{sections/11-related-work}
\input{sections/12-conclusion}
\input{sections/acknowledgment}
\bibliographystyle{ACM-Reference-Format}
\bibliography{references}
\end{document}

%% file: sections/01-introduction.tex
\section{Introduction}
\label{sec:intro}

Large language model (LLM) inference has rapidly emerged as a dominant workload in modern data centers, driving an unprecedented demand for GPU resources and electricity. Unlike traditional batch-oriented workloads, LLM serving is latency-sensitive, bursty, and increasingly deployed at scale across heterogeneous multi-GPU systems~\cite{yu2022orca, kwon2023vllm, narayanan2021megatron}. As a result, operators must simultaneously meet strict performance targets while managing increasing energy costs and power constraints~\cite{mahajan2025dynamollm, chung2023zeus, barroso2019datacenter}.

Recent systems introduced batching, request scheduling, and model-parallel execution to improve LLM inference throughput and latency~\cite{yu2022orca, kwon2023vllm, shoeybi2019megatron, narayanan2021megatron, li2023alpaserve, shahout2024fast, shahout2024don}. However, these mechanisms largely assume fixed power provisioning and treat GPU power as an external constraint rather than a controllable resource. In practice, operators rely on coarse power provisioning and static power caps to remain within rack- or cluster-level limits, but lack runtime mechanisms that can precisely and rapidly trade performance for power in response to changing workload conditions. This leads to inefficient operation, as systems either over-provision power, wasting energy, or under-provision, resulting in degraded throughput and quality-of-service (QoS) violations~\cite{chung2023zeus, mahajan2025dynamollm, barroso2019datacenter}.

Data centers are increasingly becoming subject to facility-level power caps, real-time electricity pricing, and demand-response signals from the grid~\cite{radovanovic2023carbon, zhang2021hpc, acun2023carbonexplorer}. At the same time, carbon-aware computing introduces additional incentives to modulate energy consumption over time and across locations~\cite{radovanovic2023carbon, acun2023carbonexplorer}. 
As LLM inference becomes increasingly energy-intensive, improving efficiency has become as important as maximizing throughput.
These trends require inference systems to treat power not as a fixed constraint, but as a controllable resource that can be dynamically adjusted to balance performance, efficiency, and external signals.

In this paper, we argue that GPU power caps should be treated as a first-class control primitive in LLM inference runtimes. We present a power-aware runtime that jointly optimizes hardware-level power limits and software-level parameters such as batch size and parallelism. By coordinating these knobs, the system exposes a controllable trade-off between throughput, latency, and energy efficiency.


However, enabling such control is challenging. The relationship between power, performance, and efficiency in LLM inference is highly non-linear and depends on workload characteristics, batching behavior, and system bottlenecks such as compute and communication. As a result, configurations that are efficient under one workload may become suboptimal as conditions change, making static or offline tuning insufficient. Effective control, therefore, requires lightweight predictive models and fast feedback mechanisms that operate at runtime.

Our design is guided by three observations: increasing power yields diminishing returns beyond model-dependent thresholds, batch size dominates efficiency, and the optimal degree of parallelism depends on the compute–communication balance. These observations are particularly pronounced in mixture-of-experts (MoE) models~\cite{shazeer2017moe, fedus2022switch}. In MoE architectures, tokens are dynamically routed to a subset of experts, introducing significant communication alongside computation. This leads to highly variable and often communication-bound execution, where performance depends on routing patterns, load imbalance, and interconnect bandwidth. Consequently, increasing power caps may amplify communication overhead rather than useful computation, further complicating the relationship between power and performance.

To address this challenge, we design a power-aware runtime that combines offline power–performance modeling with a feedback-driven control loop. The system continuously selects configurations that meet performance targets while maximizing efficiency, and adapts to workload changes and dynamic power budgets with low overhead. Importantly, our approach is plug-and-play: it integrates into existing LLM serving frameworks (e.g., vLLM) without requiring changes to model architectures or inference APIs. Our specific contributions are as follows:

\begin{itemize}

    \item We identify and quantify previously unexploited cross-layer interactions between GPU power caps, batching, and parallelism in LLM inference, showing that independent optimization is fundamentally suboptimal.

    \item We design a closed-loop control system that jointly tunes hardware (power caps) and software (batching) knobs under QoS constraints, enabling energy-proportional LLM serving under dynamic power budgets.

    \item We implement our design in vLLM without modifying model architectures or inference APIs, demonstrating a practical and deployable approach.

    \item We show that treating power as a first-class scheduling primitive expands the achievable Pareto frontier beyond what is possible with either dynamic voltage and frequency scaling (DVFS) or batching alone, achieving up to 26.3\% efficiency improvement and 4$\times$-7$\times$ reduction in QoS violations.
\end{itemize}

Prior work on GPU power management and DVFS treats power as a low-level hardware knob, while LLM serving systems focus on batching, scheduling, and parallelism under fixed power budgets. These two layers are optimized independently. In contrast, we show that for modern LLM inference, especially MoE models, power caps fundamentally interact with batching and parallelism through compute/communication tradeoffs, creating operating regimes that cannot be captured by either layer alone. This paper introduces the first LLM serving runtime that jointly optimizes hardware power limits and software scheduling knobs under explicit QoS constraints, enabling dynamic navigation of the power–performance–efficiency space at runtime. We show that treating power as a first-class control dimension expands the achievable efficiency–performance Pareto frontier beyond what is possible with batching or DVFS alone. Across multi-GPU systems and both dense and MoE models, our runtime improves energy efficiency by up to 26.3\%, reduces QoS violations by $4\times$-$7\times$ under power constraints.



%% file: sections/03-motivation.tex
\section{Motivation}
\label{sec:motivation}

LLM inference has emerged as a sustained, high-power workload in modern data centers, requiring systems to meet strict throughput and latency targets under growing power constraints. In practice, deployments are increasingly subject to facility-level power limits, electricity price variability, and external signals such as demand response and carbon-aware operation~\cite{barroso2019datacenter,zhang2021hpc,radovanovic2023carbon, acun2023carbonexplorer}.

Despite these constraints, existing LLM serving frameworks do not explicitly incorporate power into runtime decision-making. GPU power is typically provisioned statically, leading to either over-provisioning and energy inefficiency or under-provisioning and QoS degradation. Critically, these systems lack mechanisms to dynamically adjust power consumption in response to workload conditions and performance objectives.

To understand this gap, we conduct an empirical study of LLM inference across diverse models and configurations. Our analysis reveals that the relationship between power, performance, and efficiency is highly non-linear and strongly dependent on workload characteristics. These findings motivate the need for a runtime that treats power as a first-class control dimension and jointly optimizes it with software-level parameters.

\begin{figure}[t]
  \centering
  \includegraphics[width=\linewidth]{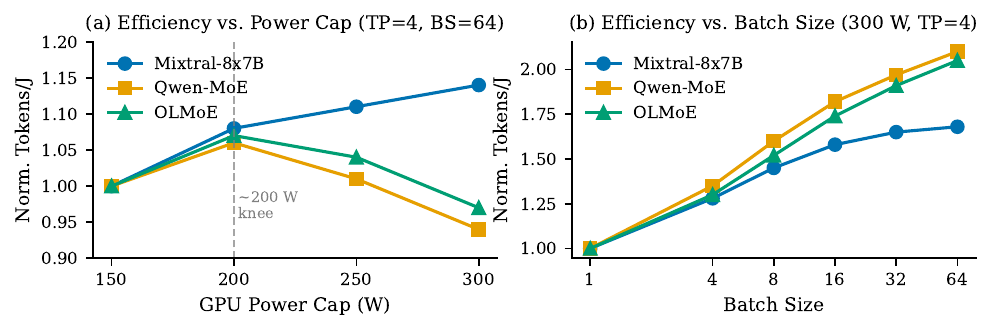}
  \vspace{-0.3in}
  \caption{(a)~tokens/J vs.\ power cap showing divergent behavior:
    compute-bound Mixtral continues to improve while communication-bound Qwen-MoE and OLMoE peak at 200\,W and decline.
    (b)~tokens/J vs.\ batch size: efficiency gains are substantial
    for all model families.}
  \label{fig:teaser}
\end{figure}

\subsection{Key Empirical Insights from Offline Profiling}


Our experiments reveal three consistent trends across models and configurations.

\paragraph{Insight 1: Power caps exhibit diminishing returns at model-dependent thresholds}
\label{sec:insight2}

Figure~\ref{fig:teaser}(a) shows tokens/J vs.\ power cap of three models: Mixtral, Qwen-MoE and OLMoE. While increasing power initially improves throughput, beyond a workload-dependent threshold (typically 150–200 W), efficiency gains diminish. This behavior varies across models depending on whether they are compute-bound or communication-bound. Compute-bound models (e.g., Mixtral) continue to benefit from higher power caps, as additional power increases SM clock frequency and compute throughput. In contrast, communication-bound models (e.g., Qwen-MoE and OLMoE) reach peak efficiency at lower power levels. Beyond this point, additional power primarily accelerates communication overhead (e.g., NCCL all-to-all traffic over NVLink) rather than useful computation, resulting in lower tokens/J. Contrary to common practice, operating GPUs at maximum power is often suboptimal for LLM inference, particularly for communication-bound workloads.

\paragraph{Insight 2: Batch size dominates power efficiency}
\label{sec:insight1}

Figure~\ref{fig:teaser}(b) shows tokens/J normalized to batch size 1.
Increasing batch size from 1 to 64 improves efficiency by
$1.7\times$--$2.1\times$ across all models.
This improvement arises from amortizing fixed per-step overheads, including kernel launches, attention computation, and communication setup, across more tokens, thereby increasing SM utilization and effective arithmetic intensity.

The magnitude of this effect depends on model characteristics. Compute-bound models (e.g., Mixtral) saturate efficiency gains at smaller batch sizes, while communication-bound models (e.g., Qwen-MoE and OLMoE) continue to benefit at larger batch sizes due to additional amortization of routing and communication overheads. For example, the marginal gain from increasing batch size from 32 to 64 is only 2\% for Mixtral, but remains 7\% for Qwen-MoE.

\paragraph{Insight 3: Compute-communication ratio determines optimal parallelism}
\label{sec:insight3}

\begin{figure}[b]
  \centering
  \includegraphics[width=\linewidth]{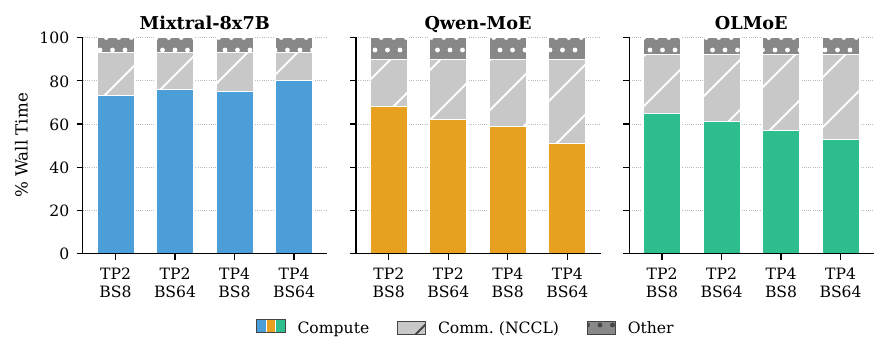}
  \vspace{-0.3in}
  \caption{Compute vs.\ communication time breakdown by model and configuration.
    Mixtral remains compute-bound; Qwen-MoE and OLMoE become
    communication-bound at higher TP and batch size.}
  \label{fig:commcomp}
\end{figure}

Figure~\ref{fig:commcomp} breaks down inference time into compute and communication components across different models and configurations. The results show that the effectiveness of power scaling depends strongly on the relative contribution of these components.

For compute-bound workloads (e.g., Mixtral under moderate parallelism), increasing GPU power caps directly improves throughput by raising SM clock frequency and accelerating arithmetic operations. As a result, higher power levels can translate into both higher throughput and improved efficiency, up to a saturation point.

In contrast, communication-bound workloads (e.g., Qwen-MoE and OLMoE under high expert parallelism) derive limited benefit from increased power. In these regimes, a significant portion of inference time is spent on communication operations such as expert routing and all-to-all exchanges over NVLink or PCIe. 

\vspace{-0.02in}


As a result, the optimal operating point shifts toward lower power caps for communication-bound configurations. Furthermore, the same power configuration can yield fundamentally different efficiency outcomes depending on the communication/computation ratio, making static configurations inherently suboptimal.

\begin{figure*}[t]
  \centering
  \includegraphics[width=\linewidth]{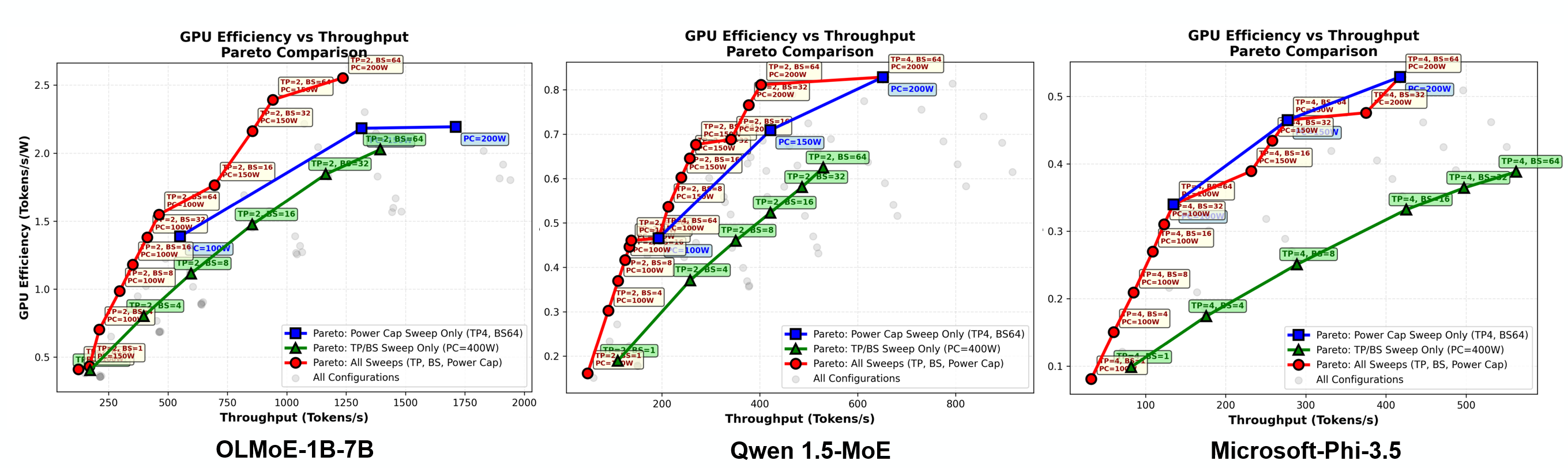}
  \vspace{-0.2in}
  \caption{Pareto frontier expansion for three MoE models (single node, 4$\times$A100).
    Four frontiers are shown: SW only (batch sweep, fixed cap),
    HW only (cap sweep, fixed batch), HW+SW (joint cap$\times$batch),
    and full joint (HW+SW+TP). The full frontier dominates any single-knob approach; gains are
    model-dependent and follow the compute/communication ratio.}
  \label{fig:pareto_expansion}
\end{figure*}

\subsection{Pareto Frontier Expansion via HW-SW Knob Combination}
\label{sec:pareto_expansion}

We quantify the benefit of jointly controlling hardware and software parameters by comparing the achievable efficiency frontier against configurations that vary only one class of knobs. Figure~\ref{fig:pareto_expansion} shows the impact of combining hardware and software control knobs on the achievable efficiency frontier. To illustrate this, we construct three Pareto frontiers for each model, each corresponding to a different set of available control knobs.

\begin{enumerate}
  \item \textbf{SW only}:  We vary batch size over $\{1,4,8,16,32,64\}$ and Tensor Parallelism (TP) over $\{1, 2, 4\}$ without applying power caps. This represents the operating range available without hardware-level power control.

  \item \textbf{HW only}: We vary the GPU power cap over its allowable range while fixing batch size and parallelism. This captures the effect of hardware-level power scaling in isolation, corresponding to what DVFS alone can achieve.

  \item \textbf{HW+SW (cap $\times$ batch $\times$ TP)}: We jointly vary power cap, batch size, and parallelism configurations, capturing the full space of available control knobs.
  
\end{enumerate}

\textbf{HW-only and SW-only frontiers are largely non-overlapping.}
Power capping alone (HW only) shifts operating points toward lower
throughput with improved efficiency, it trades performance for power.
Batch size alone (SW only) shifts toward higher throughput and higher
efficiency simultaneously, but is capped by the fixed 300\,W budget.
The two frontiers occupy different regions of the throughput-efficiency
plane and are therefore \emph{complementary}, not interchangeable.

\textbf{Combining HW+SW strictly dominates either alone.}
The joint (cap $\times$ batch $\times$ TP) frontier envelops both individual frontiers: it achieves the high-throughput end of the SW-only frontier
\emph{and} the high-efficiency end of the HW-only frontier,
with new configurations filling the gap in between.

\textbf{The expansion is model-dependent and follows the compute-communication ratio.}
For Mixtral-8x7B (compute-bound), the efficiency gain from adding
TP as a third degree of freedom is substantial. 
Higher TP at larger batch sizes unlocks operating points
that neither cap nor batch adjustment can reach alone ($+$15\% efficiency
gain at the frontier tip vs.\ HW+SW alone).
For Qwen-MoE (communication-bound), the marginal gain from TP
is smaller, TP=2 already suffices at most batch sizes
because increasing TP adds extra all-reduce overhead.
OLMoE behaves similarly to Qwen-MoE in this regard.

The full joint frontier improves peak efficiency by
$1.18\times$ (Mixtral), $1.13\times$ (Qwen-MoE), and $1.14\times$ (OLMoE)
over SW-only optimization---gains that are practically significant at
data-center scale but invisible to any system that treats HW and SW knobs
in isolation.


\begin{figure}[b]
  \centering
  \includegraphics[width=\linewidth]{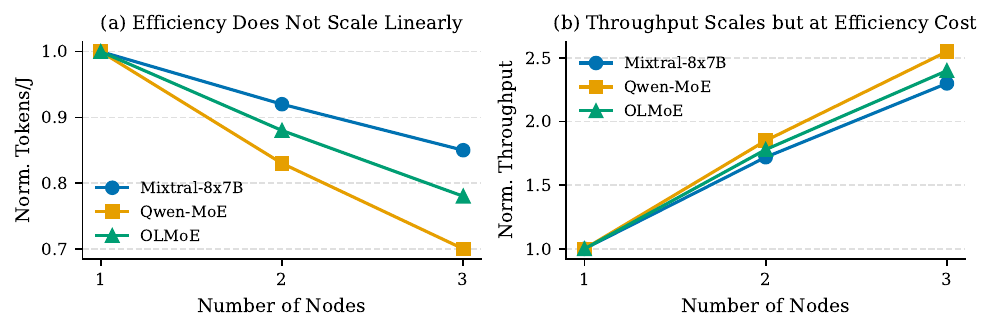}
  \vspace{-0.2in}
  \caption{Multi-node scaling: (a)~efficiency drops as node count grows,
    especially for communication-bound Qwen-MoE;
    (b)~throughput grows but at diminishing efficiency returns.}
  \label{fig:multinode}
\end{figure}

\subsection{Multi-Node Scaling Behavior}

\emph{MoE models do not always scale efficiently simply by adding more nodes.}
Figure~\ref{fig:multinode} shows normalized tokens/J and throughput
as node count increases from 1 to 3.
Qwen-MoE suffers a 30\% efficiency drop at 3 nodes because its inter-node
all-to-all traffic saturates InfiniBand bandwidth.
Mixtral scales more smoothly (15\% drop at 3 nodes)
due to higher per-expert compute density.
This means the optimal configuration strategy changes with node count and static offline policies are not sufficient.

These observations collectively reveal a fundamental limitation of existing inference systems, where no single control knob, whether batching, parallelism, or power, can achieve optimal performance and efficiency across workloads. Instead, effective operation requires coordinated control across hardware and software layers, with the ability to dynamically adapt to workload conditions and external constraints. This motivates the design of a power-aware runtime system that jointly manages GPU power caps and inference parameters to achieve energy-efficient, QoS-aware execution. 
\vspace{-0.1in}
\subsection{Impact of Expert Parallelism Across Nodes}

\begin{figure}[b]
\centering
\includegraphics[width=\linewidth]{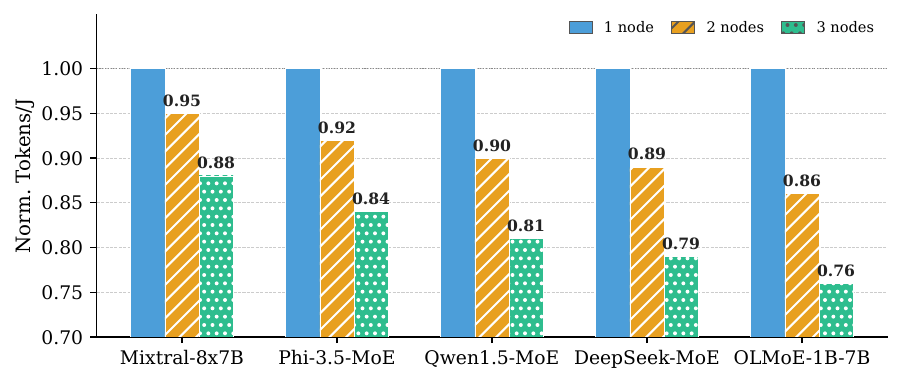}
\vspace{-0.2in}
\caption{
Energy efficiency under expert parallelism across different node counts. Each group shows normalized tokens/J 
for a given model at 1, 2, and 3 nodes. Models with higher communication intensity exhibit 
larger efficiency degradation as parallelism increases. 
}
\label{fig:ep_scaling}
\end{figure}

We evaluate the impact of expert parallelism (EP) on energy efficiency as the system scales from one to multiple nodes. 
Figure~\ref{fig:ep_scaling} shows normalized tokens per joule (tokens/J) for representative MoE models under increasing node counts.

Across all models, efficiency decreases as EP extends across nodes. This degradation is primarily driven by increased communication overhead, 
including expert routing and all-to-all exchanges, which grow with the number of participating GPUs and nodes. 
While EP improves model capacity and throughput scalability, these gains are offset by communication costs at larger scales.

The rate of efficiency degradation is strongly model dependent. Compute heavy models such as Mixtral degrade more gradually, as a larger fraction of execution time remains dominated by arithmetic operations. In contrast, communication-sensitive models such as Qwen and OLMoE experience steeper efficiency declines, reflecting their higher sensitivity to routing and synchronization overheads.


Overall, these findings confirm that scaling MoE inference through expert parallelism introduces a trade-off between capacity and efficiency. This further motivates the need for runtime mechanisms that adapt parallelism and system configuration based on workload characteristics and system constraints.

%% file: sections/new_method.tex
\vspace{-0.1in}
\section{PALS Runtime Design}
\label{sec:runtime_desing}

\begin{figure*}[t]
  \centering
  \includegraphics[width=\linewidth]{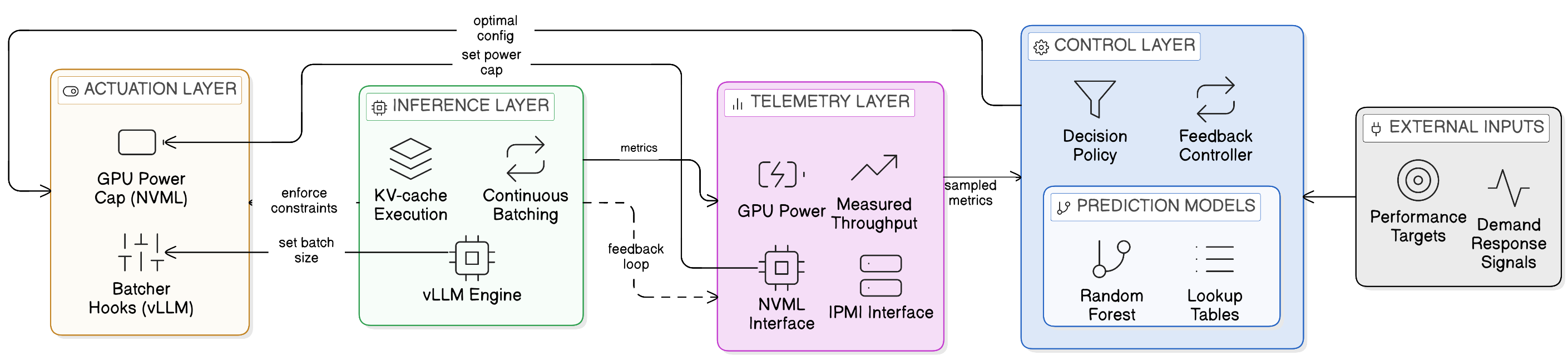}
  \vspace{-0.2in}
\caption{PALS runtime design. Telemetry from inference execution is aggregated and fed to a controller that predicts feasible operating points and issues hardware- and software-level actuation decisions.}
  \label{fig:system_architecture}
\end{figure*}

We present a power-aware runtime for LLM inference, PALS, that dynamically coordinates hardware-level power control with software-level inference parameters. It is designed to meet performance targets while maximizing energy efficiency under dynamic workload conditions and power constraints.

Figure~\ref{fig:system_architecture} illustrates the design of PALS, which consists of three key components: (1) a telemetry layer that collects real-time performance and power metrics, (2) a control layer that selects configurations based on predictive models and feedback signals, and (3) an actuation layer that enforces decisions by adjusting GPU power caps and inference parameters such as batch size.

At a high level, the runtime operates as a closed-loop controller. During each control interval, telemetry data is used to estimate current system performance. A model-guided decision module selects candidate configurations that satisfy performance targets, while a feedback controller corrects for modeling errors and runtime variability. The selected configuration is then applied through hardware and software actuation mechanisms.
This design enables the runtime to continuously adapt to workload dynamics and external signals, such as power budgets or demand-response events, while maintaining stable and efficient operation. 

\paragraph{Telemetry layer.}
The telemetry layer monitors system state in real time, including GPU power consumption, throughput (tokens/s) and GPU utilization. These signals are obtained through standard interfaces such as NVML and runtime instrumentation. The collected telemetry provides the basis for both performance estimation and feedback control.

\paragraph{Control layer.}
The control layer is responsible for selecting configurations that satisfy performance targets while optimizing energy efficiency. It combines two complementary mechanisms. First, a model-driven selection process uses lightweight power-performance model to predict the throughput and efficiency of candidate configurations. Second, a feedback controller adjusts decisions based on observed deviations from target performance, enabling robust operation under modeling inaccuracies and workload variability.

At each control interval (i.e., 500ms) , the controller evaluates feasible configurations that are defined by power caps and software-level parameters such as batch size and selects the configuration that maximizes efficiency subject to performance constraints. This process enables continuous adaptation to workload dynamics and external signals such as power budgets, performance targets or demand-response events.

\paragraph{Actuation layer.}
The actuation layer applies control decisions by interfacing with both hardware and software mechanisms. Hardware-level actuation adjusts GPU power caps using vendor-supported interfaces (e.g., NVML). Software-level actuation modifies inference parameters such as batch size and request scheduling behavior within the serving framework. These adjustments are lightweight and can be applied at runtime without interrupting ongoing inference.

\paragraph{Closed-loop operation.}
Together, these components form a closed-loop control system. Telemetry is continuously collected and used to estimate system performance, the control layer selects updated configurations, and the actuation layer enforces these decisions. This loop operates at 500ms timescale, enabling the system to respond rapidly to workload fluctuations and changing power constraints. 


This design decouples system-level decision making from the underlying inference engine, allowing our approach to be plug-and-play, integrated with existing LLM serving systems without requiring changes to model architectures or inference APIs. 

\subsection{Power–Performance Modeling}

\paragraph{Methodology}
To enable power-aware runtime optimization, we construct an offline
power-performance model from systematic profiling experiments across a wide range of inference configurations. The goal of this process is to construct a dataset that captures the mapping from inference configurations to performance and power, which is later used to train predictive models for runtime control.

We perform controlled parameter sweeps over the following dimensions:

\begin{itemize}[leftmargin=*]
  \item \textbf{Power caps.} GPU power limits are varied across the allowable
  platform range (i.e., 100-400W) using
  vendor-provided interfaces (i.e., NVML, nvidia-smi)~\cite{nvidia2024nvml, nvidia2024nvidiasmi}. This allows us to characterize
  performance scaling under various power budgets.

  \item \textbf{Batch sizes.} Continuous batching~\cite{yu2022orca, kwon2023vllm} is evaluated over a wide range (e.g., batch sizes 1-64) to capture utilization scaling effects. This range reflects realistic serving scenarios where batch sizes fluctuate with request arrival patterns.

  \item \textbf{Parallelism configurations.} We evaluate tensor parallelism (TP), expert parallelism (EP), and data parallelism (DP) where applicable. These configurations affect both compute efficiency and communication overhead, making them important factors in power-performance tradeoffs.

\end{itemize}

Table~\ref{tab:knobs} summarizes the five tunable parameters. We evaluate 
three static (require model reload) (i.e., TP, EP, DP) and two dynamic (i.e.,  batch size, power cap) knobs. During each experiment, we collect GPU power consumption (via NVML/IPMI), throughput in tokens-per-second ($\tps$), latency metrics, and system-level telemetry. This range of configurations spans both low-utilization and saturation regimes, capturing the non-linear behavior of throughput and efficiency across different operating points. This dataset serves as the foundation for the predictive models described in the next section. 

\paragraph{System-Level Power Estimation.}
We estimate system-level power from GPU telemetry by fitting a linear model
between aggregate GPU power and IPMI measurements:
$P_{\text{sys}} = \alpha \sum_i P_{\text{GPU},i} + \beta$.
Across configurations, we obtain $\alpha \approx 1.05$ and $\beta \approx 340$--$350$\,W,
with high accuracy ($R^2 \approx 0.98$, MAE 12--27\,W).
This enables reliable system-level energy estimation without continuous IPMI polling.

\begin{table}[t]
\centering\small
\caption{HW and SW control knobs. 
}
\label{tab:knobs}
\begin{tabular}{@{}lllp{2.2cm}@{}}
\toprule
\textbf{Knob} & \textbf{Cat.} & \textbf{Adj.} & \textbf{Values tested}\\
\midrule
Power Cap  & HW & Runtime & 150,200,250,300,350,400\,W \\
Batch Size & SW       & Runtime & 1,4,8,16,32,64 \\
TP         & SW       & Static  & 1, 2, 4 \\
EP         & SW       & Static  & 1, 4, 8 \\
DP         & SW       & Static  & 1, 2, 3 nodes \\
\bottomrule
\end{tabular}
\end{table}

\subsection{Prediction Model}

The prediction  model maps a configuration
\[
c = (\text{cap}, \text{batch}, \text{parallelism}, \ldots)
\]
to predicted performance and power metrics, including \tps,
average power consumption ($P$), and derived efficiency metrics such as
tokens-per-watt. The prediction model is trained offline, while the runtime policy uses it for fast decision-making with feedback correction. The model is used to score discrete candidate configurations rather than to directly control the runtime. At inference time, the controller queries the model to estimate throughput and power for feasible operating points and ranks them according to the optimization objective.

We use a lightweight regression model to predict throughput and power as a function of configuration parameters. We evalaute various machine learning algorithms, such as XGBoost, gradient boosting and random forest. We empirically determine that random forest provide the most accurate for both MoE and dense models. Random forest model is able to capture non-linear relationships between configuration parameters and system performance while maintaining low inference overhead. These relationships arise from interactions between batching, power scaling, and communication overhead, which vary across model architectures and parallelism configurations. The model provides fast, approximate predictions of performance and power for candidate configurations, enabling efficient exploration of the configuration space at runtime. We evaluate the accuracy of the offline prediction model using held-out configurations. Across all workloads, the random forest model achieves a mean absolute percentage error (MAPE) of 6.8\% for throughput prediction and 4.5\% for power prediction. Furthermore, feature importance analysis shows that batch size is the dominant predictor of efficiency, followed by power cap and tensor parallelism degree.
This aligns with our empirical observations in Section~\ref{sec:motivation}, where batching provides the largest gains.





This modeling approach provides sufficiently accurate estimates for candidate ranking while avoiding costly online exploration. While the offline predictor provides fast estimates of throughput and power, residual error and workload variability are handled by the online controller described in the next section.





\subsection{Runtime Control Policy}


At runtime, the controller uses the offline prediction model to evaluate feasible configurations and select operating points that satisfy performance constraints while maximizing energy efficiency.

\paragraph{Control Problem.}
We formulate power-aware LLM inference as a constrained optimization problem over a discrete configuration space. At each control interval, the runtime selects a configuration, $c$, 
that determines both hardware-level power limits and software-level execution parameters.

The objective is to maximize energy efficiency, measured in tokens per joule, while satisfying performance constraints. Formally, we seek to solve:
\[
\max_{c \in C(t)} \mathrm{Eff}(c)
\quad \text{s.t.} \quad
T(c) \ge T_{\text{target}} .
\]
where $C(t)$ denotes the set of feasible configurations at control interval $t$. In practice, system dynamics and workload variability make it difficult to directly solve this problem online. The mapping from configuration to performance is non-linear and workload-dependent, and optimal configurations may shift over time as request patterns and system conditions change.

These challenges motivate a model-guided control approach that combines offline prediction with online feedback correction. The offline model provides fast estimates of throughput and power for candidate configurations, while the feedback controller compensates for prediction error and workload variability during runtime.

\begin{algorithm}[t]
\caption{PALS Runtime Control}
\label{alg:control}
\begin{algorithmic}[1]

\State \textbf{Input:} $C$, $T_{\text{target}}$, $\epsilon$

\While{every control interval}

    \State Measure throughput $T_{\text{meas}}$
    \State $e \gets T_{\text{target}} - T_{\text{meas}}$

    \State $C_t \gets$ feasible configurations

    \ForAll{$c \in C_t$}
        \State $(T(c), P(c)) \gets \text{Model}(c)$
    \EndFor

    \State $C_{\text{ok}} \gets \{c \mid T(c) \ge T_{\text{target}}\}$

    \If{$C_{\text{ok}} \neq \emptyset$}
        \State $c^* \gets \arg\max_{c \in C_{\text{ok}}} \frac{T(c)}{P(c)}$
    \Else
        \State $c^* \gets \arg\max_{c \in C_t} T(c)$
    \EndIf

    \If{$|e| > \epsilon$ \textbf{for sustained intervals}}
        \State Apply $c^*$
    \EndIf

\EndWhile

\end{algorithmic}
\end{algorithm}

\paragraph{Decision Policy.}
To approximately solve the control problem online, the runtime combines model-guided candidate selection with feedback-driven correction. Algorithm~\ref{alg:control} summarizes the runtime decision loop. At each control interval, the controller enumerates feasible configurations and queries the offline model to estimate throughput and power. It then constructs the subset $C_{\mathrm{ok}}$ of configurations that satisfy the throughput target. If $C_{\mathrm{ok}} \neq \emptyset$, the controller selects the configuration with the highest predicted efficiency; otherwise, it falls back to the configuration with the highest predicted throughput. 


\paragraph{Model-guided selection.}
At each control interval, the controller evaluates feasible candidate configurations using the offline predictor and discards those that do not satisfy the throughput target. Among the remaining candidates, it selects the configuration with the highest predicted efficiency.

\paragraph{Feedback adjustment.}
As the offline predictor is approximate, the runtime continuously compares predicted behavior against measured throughput and applies correction based on throughput error using a PID controller. This improves robustness to modeling error and workload variability by compensating for steady-state mismatch and transient deviations. To avoid oscillations, configuration updates are applied only when deviations persist beyond a threshold (5\%) or when external constraints, such as power budgets or throughput targets, change. This prevents frequent reconfiguration in response to transient fluctuations.

%% file: sections/08-evaluation.tex
\section{Evaluation Methodology}
\label{sec:eval}

\subsection{Experimental Setup}

We evaluate PALS on multi-GPU server nodes equipped with NVIDIA A100 GPUs. Each node consists of 4×A100 GPUs interconnected via NVLink. This setup allows us to evaluate both compute-bound and communication-bound inference regimes.

We evaluate both dense and MoE models to capture a range of compute–communication characteristics. Experiments are conducted under both steady-state and dynamic conditions, including scenarios with time-varying power budgets. 

Workloads consist of request streams with configurable arrival rates and sequence lengths, simulating realistic inference serving conditions. We use Poisson arrival processes to model bursty request patterns commonly observed in production LLM services.

Our implementation is built on top of the vLLM inference framework~\cite{kwon2023vllm}. We extend the runtime scheduler to expose batch size as a dynamically controllable parameter, while GPU power caps are enforced through NVML interfaces. The control loop operates externally to the core execution engine and periodically updates runtime parameters without modifying model execution logic. We evaluate the system using standard benchmark datasets, including HellaSwag and GSM8K~\cite{zellers2019hellaswag, cobbe2021gsm8k}, to generate realistic inference workloads with diverse sequence lengths and reasoning complexity.

\subsection{Workloads}
We evaluate PALS across a diverse set of LLMs, including both dense and MoE architectures. Table~\ref{tab:workloads} summarizes the evaluated models, including total parameter count, active parameters per token, and parallelism characteristics. The workload set includes representative models including GPT-2~\cite{radford2019gpt2}, Llama-2-7B~\cite{touvron2023llama2}, Mistral-7B~\cite{jiang2023mistral}, Mixtral--8x7B~\cite{jiang2024mixtral}, Qwen1.5-MoE~\cite{qwen2024moe}, OLMoE-1B-7B~\cite{muennighoff2024olmoe}, DeepSeek-MoE~\cite{dai2024deepseekmoe}, and
Phi-3.5-MoE~\cite{abdin2024phi3}. These models span a wide range of compute and communication behaviors, enabling evaluation under both compute-bound and communication-bound regimes.

\begin{table}[t]
\centering
\small
\setlength{\tabcolsep}{3pt}
\begin{tabular}{lrrrrl}
\toprule
\textbf{Model} & \textbf{Tot} & \textbf{Act} & \textbf{Exp} & \textbf{$k$} & \textbf{Type} \\
               & (B) & (B) &  &  &  \\
\midrule
GPT-2        & 0.12 & 0.12 & -- & -- & Dense \\
Llama-2-7B   & 6.74 & 6.74 & -- & -- & Dense \\
Mistral-7B   & 7.25 & 7.25 & -- & -- & Dense \\
\midrule
OLMoE-1B-7B  & 6.92 & 1.30 & 64 & 8  & MoE \\
Qwen1.5-MoE$^\ddagger$ & 14.3 & 2.70 & 64 & 4  & MoE \\
DeepSeek-MoE$^\dagger$ & 16.4 & 2.80 & 66 & 6  & MoE \\
Phi-3.5-MoE  & 41.9 & 6.60 & 16 & 2  & MoE \\
Mixtral-8x7B & 46.7 & 12.9 & 8  & 2  & MoE \\
\bottomrule
\end{tabular}
\caption{
Model summary. \textbf{Tot} and \textbf{Act} denote total and active parameters (in billions).
For MoE models, \textbf{Exp} is the number of experts per layer and $k$ is the number of selected experts (top-$k$).
Dense models activate all parameters ($\text{Tot}=\text{Act}$), while MoE models activate only a subset, leading to lower effective compute despite larger total size.
}
\label{tab:workloads}
\end{table}

\subsection{Baselines}
We compare PALS against four representative control strategies that isolate different dimensions of runtime adaptation. The \emph{Baseline} uses a fixed 400W power cap and maximum batch size, reflecting a throughput-optimized production configuration without dynamic adaptation. \emph{Adaptive Batch} keeps the power cap fixed while selecting batch size using the offline prediction model, capturing software-level optimization in isolation. \emph{Adaptive Cap} dynamically adjusts GPU power caps based on QoS requirements while keeping batch size fixed, representing hardware-level control without coordination with batching. In contrast, \emph{PALS} jointly adapts both power caps and batch size using model-guided selection combined with feedback control. Finally, we include an \emph{Oracle} that performs an exhaustive offline search over all configurations, providing an upper bound on achievable efficiency. All methods are implemented within the same vLLM serving framework, only the control policy differs.

\subsection{Metrics}
We evaluate system performance using a set of complementary metrics that capture both service quality and efficiency. Throughput, measured in tokens per second, reflects the overall inference performance of the system. To quantify service reliability, we measure the QoS violation rate, defined as the fraction of time the system fails to meet target throughput thresholds. Under constrained operation, we also evaluate power tracking error, which measures the deviation between actual and target power consumption. Finally, we assess energy efficiency in terms of tokens per joule, enabling comparison across configurations with different power budgets and capturing the effectiveness of joint power–performance optimization.

%% file: sections/09-results.tex
\section{Results}
\label{sec:results}

We evaluate PALS along three dimensions: (1) single-node efficiency, (2) QoS under power-constrained multi-node operation, and (3) responsiveness to dynamic power signals in demand-response scenarios. These experiments validate that jointly controlling hardware (power caps) and software (batching) expands the achievable operating space and enables robust performance under power constraints.

\subsection{Single-Node Efficiency}
\label{sec:eval_single}
\begin{figure}[b]
  \centering
  \includegraphics[width=0.92\linewidth]{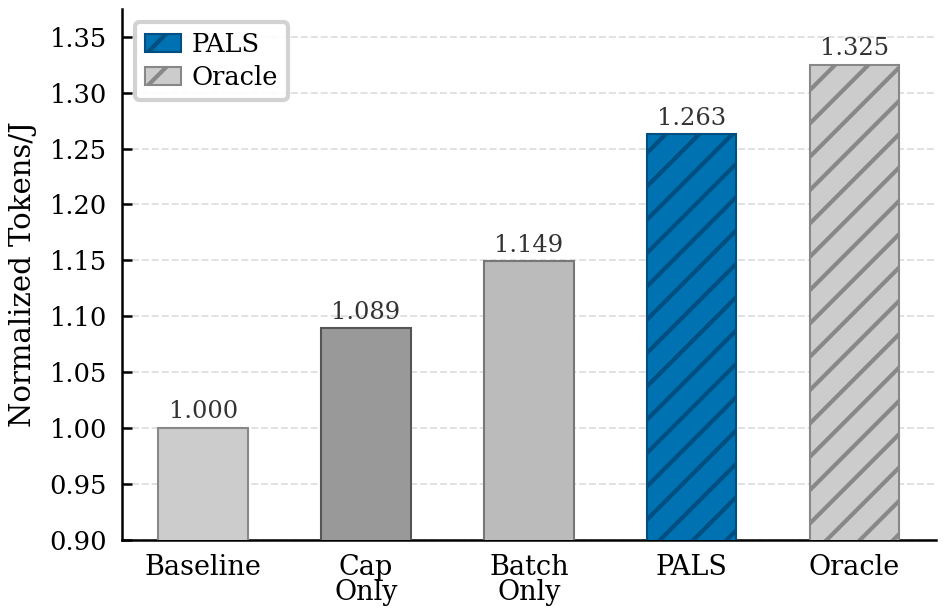}
  \vspace{-0.2in}
  \caption{Normalized tokens/J, average of five MoE models and three dense models.
    \sys achieves 26.3\% improvement over baseline and reaches 95\% of oracle efficiency.}
  \label{fig:efficiency}
\end{figure}

We first evaluate whether joint control of power caps and batching improves energy efficiency on a single node. Figure~\ref{fig:efficiency} compares five configurations on five MoE models in Table~\ref{tab:workloads}: (1) Baseline (400W, BS=64), (2) Cap Only (model-selected power cap, fixed batch), (3) Batch Only (maximum power cap, model-selected batch size), (4) PALS (jointly optimized power cap and batch size), and (5) Oracle (exhaustive offline optimum).

PALS achieves 1.26× normalized efficiency, capturing 95\% of the oracle headroom. Batch-only adaptation (1.14×) accounts for most of the improvement, indicating that batching is the dominant factor in efficiency. Power cap adaptation provides an additional gain when combined with batching.

This behavior reflects the Pareto frontier expansion observed in Section~\ref{sec:pareto_expansion}, where jointly controlling hardware and software knobs enables operating points that are not achievable with either knob alone. Consistent with our empirical observations, batching improves utilization by amortizing overheads, while power caps exhibit diminishing returns beyond model-dependent thresholds.

Overall, these results show that while batching drives most efficiency gains, power control provides a complementary benefit, and joint optimization is necessary to approach near-optimal operating points.

\subsubsection{Near-Optimality Under Runtime-Feasible Knobs}

\begin{figure}[t]
    \centering
    \includegraphics[width=0.9\linewidth]{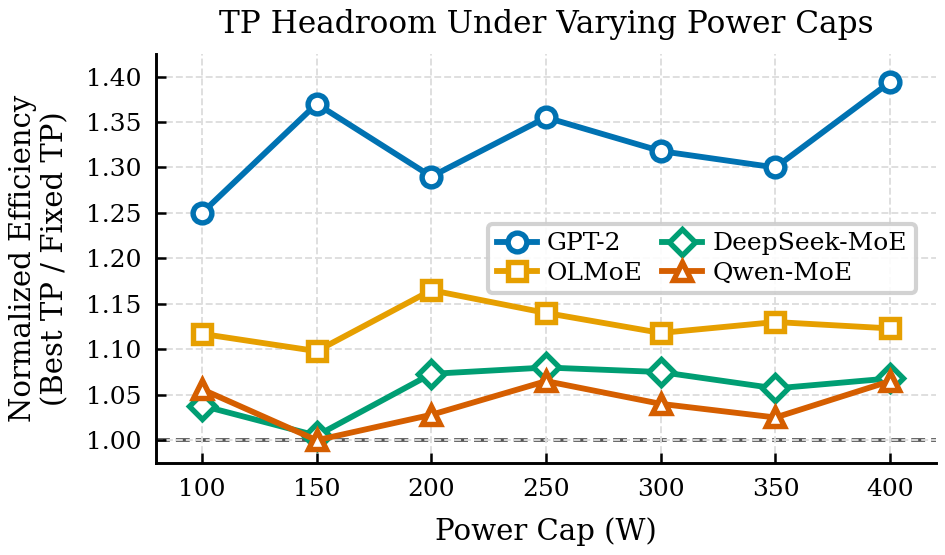}
    \vspace{-0.15in}
    \caption{Efficiency headroom from TP under varying power caps. Each curve shows the gap between a fixed-TP deployment and the best offline TP choice. Headroom varies with changing power caps, indicating that the benefit of TP selection is power-dependent.}
    \label{fig:tp_headroom_overlay}
\end{figure}

PALS adapts only runtime-feasible knobs, namely GPU power caps and batch size,
while tensor parallelism (TP) is fixed at deployment time. To quantify the resulting gap, we compare a fixed-TP configuration against the best offline TP choice at each power cap. Figure~\ref{fig:tp_headroom_overlay} shows the efficiency headroom for models with TP sensitivity. For many models, fixed TP achieves near-optimal efficiency across all power caps, indicating that runtime adaptation alone captures most gains. However, for a subset of models, the gap widens under lower power caps. This trend reflects a shift toward communication-sensitive regimes under constrained power, where the optimal compute-communication balance depends on TP. Overall, PALS achieves near-optimal efficiency using only runtime-feasible knobs, with remaining headroom attributable to deployment-time parallelism choices.

\subsection{QoS Under Multi-node Power Constraints}
\label{sec:eval_qos}
\begin{figure}[t]
  \centering
  \includegraphics[width=\linewidth]{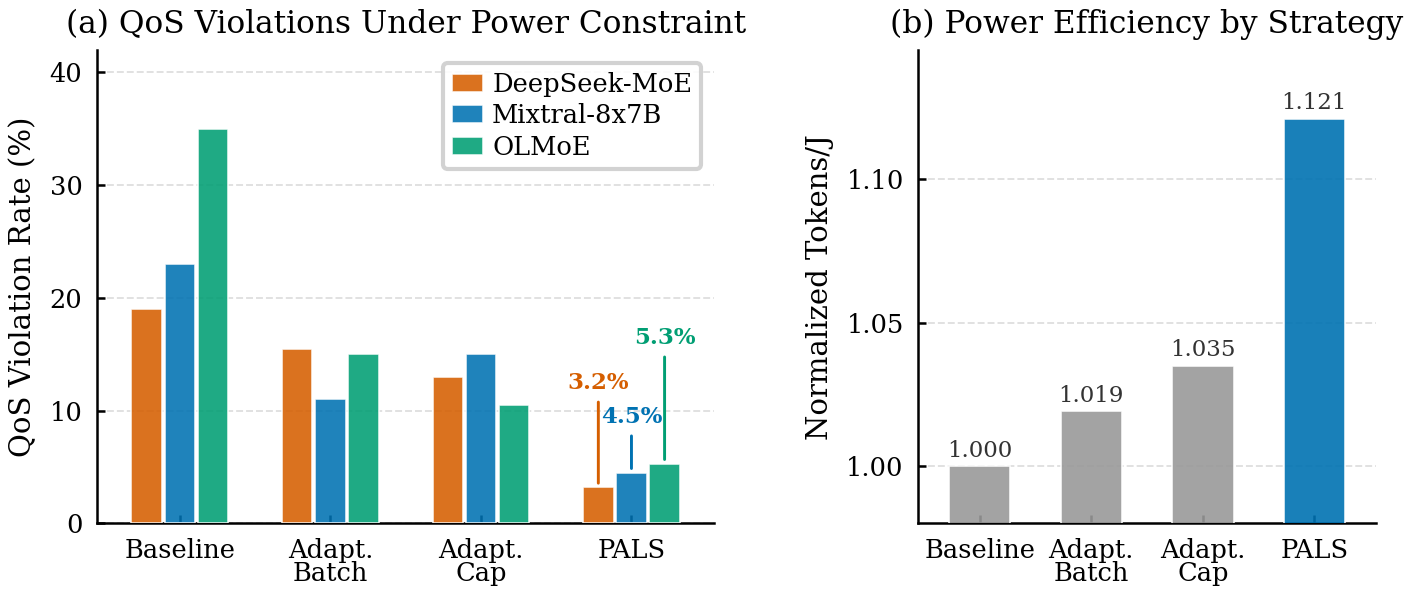}
  \caption{Multi-node power-constrained evaluation (3 nodes, 60 min).
    (a)~QoS violation rates: \sys reduces violations by 4$\times$--7$\times$.
    (b)~Normalized aggregate efficiency by strategy.}
  \label{fig:qos}
\end{figure}

We next evaluate whether PALS maintains QoS under constrained power budgets in a multi-node setting. We consider a three-node deployment, where each node runs a different model (e.g., DeepSeek-MoE, Mixtral-8x7B, and OLMoE) under a maximum cluster-wide power budget of 4{,}800W. Each model is assigned a QoS constraint, defined as a target throughput relative to its unconstrained performance (e.g., 90\%, 60\%, and 75\% for DeepSeek-MoE, Mixtral, and OLMoE, respectively). Workloads follow a Poisson arrival process over 60-minute runs. We compare four strategies corresponding to different levels of control: no adaptation (Baseline), software-only adaptation (Adaptive Batch), hardware-only adaptation (Adaptive Cap), and joint hardware–software optimization (PALS). Figure~\ref{fig:qos}(a) shows QoS violation rates. The baseline exhibits high violation rates (18.7\%-35.2\%), reflecting its inability to adapt under constrained power. Adaptive batch and adaptive cap reduce violations modestly, but remain insufficient.

In contrast, PALS reduces QoS violations to 3.2\%-5.3\%, achieving a 4×–7× improvement across all models. At the same time, it improves aggregate energy efficiency by 12.1\% (Figure~\ref{fig:qos}(b)). Adaptive batch alone cannot compensate when models become power-constrained, while adaptive power capping assumes a simplified relationship between power and performance that does not hold in practice. These limitations explain the gap between partial adaptation strategies and the full joint controller.

Overall, these results demonstrate that the key benefit of PALS is not simply adapting to power constraints, but maintaining QoS while doing so. Joint optimization of power and batching is required to simultaneously satisfy QoS targets and achieve efficient operation under constrained power budgets.

\begin{figure}[b]
  \centering
  \includegraphics[width=\linewidth]{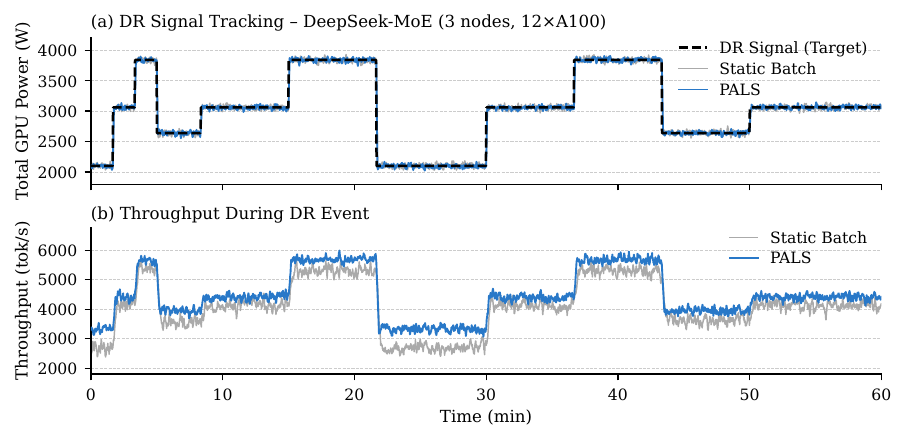}
  \vspace{-0.2in}
  \caption{Grid demand-response tracking (1-hour, DeepSeek-MoE, 3-nodes).  \sys maintains higher throughput at low power targets by co-adapting batch size. PALS improves throughput by up to 22\% at low power targets compared to the static-batch baseline.}
  \label{fig:grid}
\end{figure}

\subsection{Grid-Interactive Demand Response}
\label{sec:eval_grid}

We evaluate whether PALS can operate as a grid-interactive workload by dynamically adapting to external power signals. We simulate a demand-response (DR) scenario over 1-hour on a 3-node system running DeepSeek-MoE. We compare a static-batch baseline (i.e., fixed BS=64, dynamic cap only) against PALS, which jointly adapts power cap and batch size. 

Figure~\ref{fig:grid}(a) shows that both approaches are able to follow the target power signal. However, Figure~\ref{fig:grid}(b) shows a key difference: PALS maintains significantly higher throughput at lower power targets by dynamically adjusting batch size to match available compute capacity. In contrast, the static baseline, while able to track the power signal, suffers from underutilization at low power levels, as large batch sizes become inefficient under constrained power. PALS improves throughput by up to 22\% at low power targets compared to the static-batch baseline.

\begin{figure}[t]
  \centering
  \includegraphics[width=\linewidth]{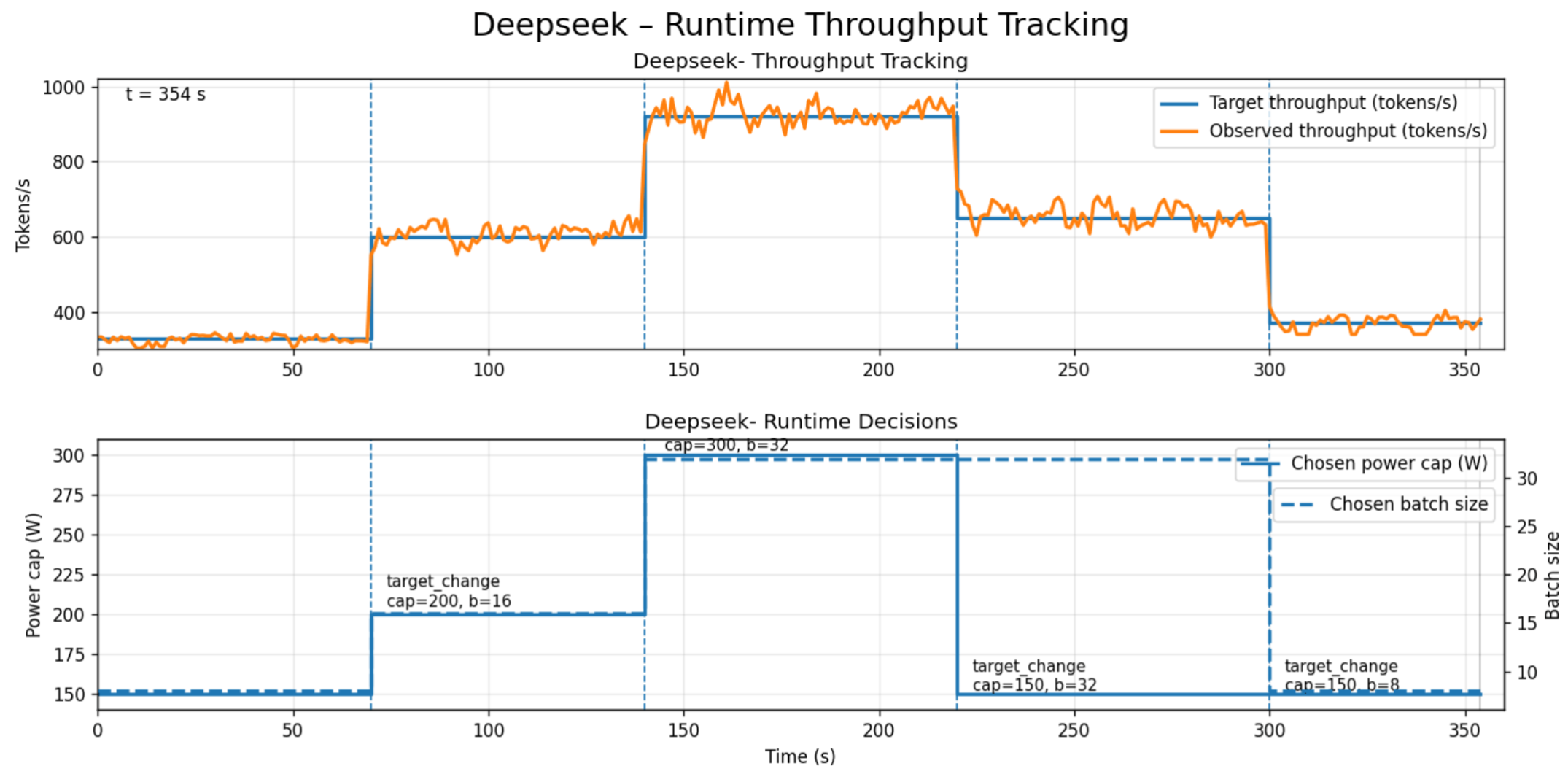}
  \caption{PALS is able to track throughput target by dynamically tuning batch size and power cap decisions at runtie.}
  \label{fig:throughput}
\end{figure}

%% file: sections/10-discussion.tex
\section{Discussion}
\label{sec:discussion}

PALS' predictive models are built from offline profiling over a fixed set of configurations and a representative set of prompts. When the online workload deviates
significantly, for example, due to unusually long prompts,
adversarial inputs, or shifts in output length distribution, the offline model may mispredict throughput or efficiency, leading to suboptimal configuration selection. The PID
feedback controller partially compensates for such errors by correcting steady-state deviations, but it cannot anticipate sudden load changes. Incorporating lightweight online model updates or uncertainty-aware prediction could reduce this gap.

PALS operates at the node level and is designed to complement, not replace, cluster-level energy management systems such as DynamoLLM~\cite{dynamo2025}. Cluster-level systems are well-suited to coarse-grained decisions, such as instance counts, routing policies, and aggregate power budgets, but cannot react to sub-second workload fluctuations within a single serving node. Our solution fills this gap, as the main scheduler sets per-node power budgets and throughput targets as external signals, which PA-vLLM tracks with its closed-loop controller. This division of responsibility, the cluster decides \emph{where and how
much}, the node runtime decides \emph{how} enables coordinated control across timescales without requiring either layer to solve the full joint optimization problem, providing a practical path toward grid-interactive LLM serving at scale.

%% file: sections/11-related-work.tex
\section{Related Work}
\label{sec:related}

\begin{table*}[t]
\setlength{\tabcolsep}{4pt}
\centering\small
\caption{Comparison with closely related systems.}
\label{tab:related}

\begin{tabularx}{\textwidth}{@{}lXXc cX@{}}
\toprule
\textbf{System} & \textbf{Scope} & \textbf{Objective}
  & \textbf{Knobs} & \textbf{MoE} & \textbf{Key result}\\
\midrule
\textbf{\sys (this work)} &
  Node/multi-node inference &
  Max tokens/J under power + QoS &
  Power cap + batch (runtime) &
  \checkmark &
  +26.3\% eff.; $4\times$--$7\times$ QoS improvement \\[2pt]

DynamoLLM~\cite{dynamo2025} &
  Cluster routing + instance mgmt &
  Min energy/cost/carbon w/ SLOs &
  Routing, GPU freq, instance cnt &
  $\times$ &
  52\% energy; 38\% carbon savings \\[2pt]

TAPAS~\cite{tapas2025} &
  VM placement in GPU clusters &
  Reduce TCO via cooling/power &
  VM placement, routing &
  $\times$ &
  Thermal/power throttling reduction \\[2pt]

Zeus~\cite{chung2023zeus} &
  Single-GPU DNN training &
  Energy-time trade-off &
  Power limit + batch (offline) &
  $\times$ &
  15-76\% training energy savings \\
\bottomrule
\end{tabularx}
\end{table*}
\textbf{LLM inference serving systems.}
A substantial line of work focus on improving LLM inference performance by
optimizing scheduling, memory management, and parallelism under fixed
power budgets.
Orca~\cite{yu2022orca} introduce iteration-level scheduling and
selective batching, enabling continuous batching that keeps GPUs
saturated across diverse request streams.
vLLM~\cite{kwon2023vllm} propose a technique to eleminate KV-cache fragmentation via
PagedAttention, allowing larger effective batch sizes and
2-4$\times$ throughput gains over static-batch systems.
Sarathi-Serve~\cite{agrawal2024sarathi} further decoupled prefill
and decode latency through chunked prefills and stall-free scheduling.
DistServe~\cite{zhong2024distserve} disaggregates prefill and decode
onto separate GPU pools, independently optimizing time-to-first-token
and per-token generation latency.
AlpaServe~\cite{li2023alpaserve} uses statistical multiplexing with
model parallelism to improve GPU utilization across concurrent
bursty workloads.
FlexGen~\cite{sheng2023flexgen} maximizes throughput under tight
memory constraints by orchestrating offloading across GPU, CPU, and
disk via a linear-programming policy.
Helix~\cite{mei2025helix} serves LLMs on heterogeneous GPU clusters
by formulating placement and request routing as a max-flow problem.
All of these systems treat GPU power as a fixed constraint; PA-vLLM
introduces power caps as a first-class scheduling dimension that
expands the achievable efficiency frontier beyond software-only
optimization.
 
\textbf{Model parallelism for large-scale LLM execution.}
Megatron-LM~\cite{shoeybi2019megatron, narayanan2021megatron} defines
the standard composition of tensor, pipeline, and data parallelism for
training and serving multi-billion-parameter models on multi-GPU
clusters. These works show that parallelism configurations critically affect
the compute-communication balance, a relationship PA-vLLM directly
exploits: our profiling reveals that the optimal power cap depends
on how compute- or communication-bound a given parallelism
configuration renders a model, particularly in MoE architectures.
 
\textbf{Mixture-of-Experts inference.}
Sparsely-gated MoE~\cite{shazeer2017moe} introduce conditional
computation via learned routing, scaling model capacity without
proportional increases in active parameters.
Switch Transformers~\cite{fedus2022switch} scales this approach to
trillion-parameter regimes, highlighting load-balancing and
communication as primary system challenges.
GShard~\cite{lepikhin2020gshard} targets distributed MoE training
with expert parallelism across accelerators, and
DeepSpeed-Inference~\cite{aminabadi2022deepspeed} optimizes MoE
inference at scale via kernel fusion and communication scheduling.
Expert offloading~\cite{wu2024expertoffloading} reduces memory
pressure by paging inactive experts to CPU.

\textbf{GPU power management and DVFS for ML workloads.}
Dynamic voltage and frequency scaling (DVFS) has long been studied
as a hardware knob for energy-performance tradeoffs in
processors~\cite{barroso2019datacenter}.
Zeus~\cite{chung2023zeus} applies joint power-limit and batch-size
optimization to DNN training on a single GPU, achieving 15-76\%
energy savings through offline exploration.
Nabavinejad et al.~\cite{nabavinejad2022dvfs} propose BatchDVFS, a runtime system that coordinates batch size and DVFS to meet power caps in single-GPU DNN inference, demonstrating that the achievable power range via DVFS alone depends on batch-determined utilization. PA-vLLM extends this cross-layer insight to multi-GPU LLM serving, where continuous batching, autoregressive generation, and MoE communication overhead create fundamentally different power-performance dynamics that batch-size and frequency tuning in isolation cannot capture. Unlike BatchDVFS, PA-vLLM incorporates parallelism as a third control dimension, operates under a closed-loop feedback controller that responds to dynamic external power budgets, and integrates into a production serving framework targeting QoS-constrained inference. Patel et al.~\cite{patel2024llmpower}
characterize LLM power consumption patterns in production cloud clusters
and proposed POLCA, a power oversubscription framework that
safely deploys 30\% more inference servers within fixed datacenter
power budgets.
PA-vLLM is complementary to POLCA, where POLCA targets
cluster-level provisioning, PA-vLLM operates within a single
inference node and dynamically co-adapts per-GPU power caps and
batch size at 500ms granularity to maintain QoS while maximizing
tokens per joule.
 
\textbf{Cluster-level LLM energy management.}
DynamoLLM reduces energy and carbon cost
at the cluster level by dynamically reconfiguring instance counts,
routing policies, and GPU frequencies to meet latency SLOs, however
operates at the granularity of minutes and does not address
sub-second variability within a running inference instance~\cite{mahajan2025dynamollm} .
TAPAS~\cite{bhoria2025tapas} reduces total cost of ownership through
thermal- and power-aware VM placement in GPU clusters, treating power as an infrastructure-level capacity constraint.
PA-vLLM operates at the node level and complements both systems:
a cluster scheduler sets per-node budgets and throughput targets,
which PA-vLLM tracks with its local controller, enabling
coordinated control across timescales without either layer solving
the full joint optimization.

\paragraph{Distinguishing characteristics of Power-Aware vLLM}
Compared to prior work, Power-Aware vLLM makes three key departures. First, it elevates GPU power caps to a first-class runtime control primitive and integrates them directly into the inference scheduling loop, rather than treating power as a static provisioning constraint. Second, it explicitly models the interaction between power, batching, and the communication/computation ratio of LLM inference, enabling stable control even in communication-bound regimes common in multi-GPU and MoE deployments. Third, it targets production LLM serving scenarios, requiring sub-second response times, negligible overhead, and zero changes to model architectures or inference APIs.

Together, these differences position Power-Aware vLLM as the first system to provide fine-grained, communication-aware power control for multi-GPU LLM inference at runtime, complementing existing job-level, cluster-level, and data center–level energy management solutions.

%% file: sections/12-conclusion.tex
\section{Conclusion}
\label{sec:conclusion}

In this paper, we present a plug-and-play power-aware runtime that treats GPU power caps as a first-class control knob, jointly optimizing them with batch size and parallelism inside an unmodified LLM engine deployment. Our characterization reveals that power efficiency exhibits model-dependent diminishing returns, that batch size dominates tokens per joule, and that the compute-communication ratio determines whether additional power aids computation or amplifies overhead, particularly in MoE models. Jointly controlling hardware and software knobs expands the achievable efficiency frontier beyond what either alone can reach.
 
We implement our approach within vLLM. Across single- and multi-node evaluations on dense and MoE models, our runtime improves energy efficiency by up to 26.3\%, reduces QoS violations by 4$\times$--7$\times$ under multi-node power budgets, and tracks dynamic demand-response signals without requiring changes to model architectures or inference APIs. These results establish GPU power as a practical runtime control dimension for LLM inference, opening a path toward energy-proportional, grid-interactive AI serving at scale.
 

%% file: sections/acknowledgment.tex
\section{Acknowledgments}
Authors used generative AI tools to assist with light editing and refinement of the text. 